\documentclass[square,sort,comma,10pt,twocolumn,letterpaper]{article}
\pdfoutput=1 
\usepackage[accsupp]{axessibility}  \pdfoutput=1 

\usepackage{iccv}
\usepackage{times}
\usepackage{epsfig}
\usepackage{graphicx}
\usepackage{amsmath}
\usepackage{amssymb}

\usepackage{microtype}
\usepackage{subfigure}
\usepackage{booktabs} \usepackage{soul,color}
\usepackage{multirow}
\usepackage{enumitem}
\usepackage{algorithm}
\usepackage{algorithmic}
\usepackage{pdfpages}
\usepackage{mathtools}

\usepackage[pagebackref=true,breaklinks=true,letterpaper=true,colorlinks,bookmarks=false]{hyperref}

\iccvfinalcopy 

\def\iccvPaperID{7961} 

\ificcvfinal\fi

\begin{document}

\title{Learning Fast Sample Re-weighting Without Reward Data}

\author{Zizhao Zhang \quad Tomas Pfister \\
Google Cloud AI\\
}

\maketitle
\ificcvfinal\thispagestyle{empty}\fi

\begin{abstract}
Training sample re-weighting is an effective approach for tackling data biases such as imbalanced and corrupted labels. 
Recent methods develop learning-based algorithms to learn sample re-weighting strategies jointly with model training based on the frameworks of reinforcement learning and meta learning. 
However, depending on additional unbiased reward data is limiting their general applicability. 
Furthermore, existing learning-based sample re-weighting methods require nested optimizations of models and weighting parameters, which requires expensive second-order computation. 
This paper addresses these two problems and presents a novel learning-based fast sample re-weighting (FSR) method that does not require additional reward data. 
The method is based on two key ideas: learning from history to build proxy reward data and feature sharing to reduce the optimization cost. 
Our experiments show the proposed method achieves competitive results compared to state of the arts on label noise robustness and long-tailed recognition, and does so while achieving significantly improved training efficiency. The source code is publicly~available at~\url{https://github.com/google-research/google-research/tree/master/ieg}.
\end{abstract}

\section{Introduction}
The performance of DNNs is dependent on the scale of training datasets and the quality of labels. 
Data biases are inevitable in practice, and in particular, noisy labels \cite{tanaka2018joint,zhang2020distilling} or imbalanced classes \cite{cui2019class,kang2019decoupling} can negatively influence the model performance.

Sample re-weighting is an effective strategy that has been explored to address problems caused by data biases \cite{byrd2019effect}. 
The underline principle of sample re-weighting is as simple as upgrading the weights of good samples and downgrading the weights of bad samples. 
Finding effective weights that can optimize the model training with stochastic gradient descent (SGD) is a dynamic process. 
The optimal weight of a sample can change over model training phases. 
Over-weighting simple samples at later phases while under-weighting hard samples at early phases can cause negative effects to the overall DNN accuracy \cite{hacohen2019power}. 
In light of the advances in meta learning and reinforcement learning (RL), there is a growing interest in learning-to-learn based algorithms to optimize sample weights jointly with model training \cite{ren2018learning,shu2019meta,hu2019learning,yoon2020data}. 
This problem well resembles the design of MAML~\cite{finn2017model}: incorporating meta optimization inside the supervised training for sample weight optimization.
The meta-objective of re-weighting is usually defined as finding the optimal weight per sample such that the trained model has the best objective on an additional reward (a.k.a. validation) dataset. 
Moreover, such reward dataset is required to be unbiased and to have reasonable size.
For example, in label noise robust training problems, this unbiased reward dataset is expected to have clean and class-balanced labels. 
This extra requirement has been noted to be problematic, but how to remove such a requirement remains unanswered \cite{cao2019learning}.

From an optimization and efficiency perspective, although this problem can be directly formulated as an RL problem, the training computation is very expensive \cite{ghorbani2019data,yoon2020data}. 
Therefore, most existing work follows the more efficient meta learning framework, assuming that the weight optimization with respect to reward signals is a fully differentiable problem. 
Even so, similarly to MAML, the overall computation still requires a second-order unroll of DNN computation graphs, which increases the memory requirement and training time complexity significantly \cite{ren2018learning}. 
Such a limitation hinders the applicability of the method to large-scale DNNs, and emphasizes the importance of the need for further improving efficiency.

\begin{figure}[t]
     \centering
     \includegraphics[width=0.99\linewidth]{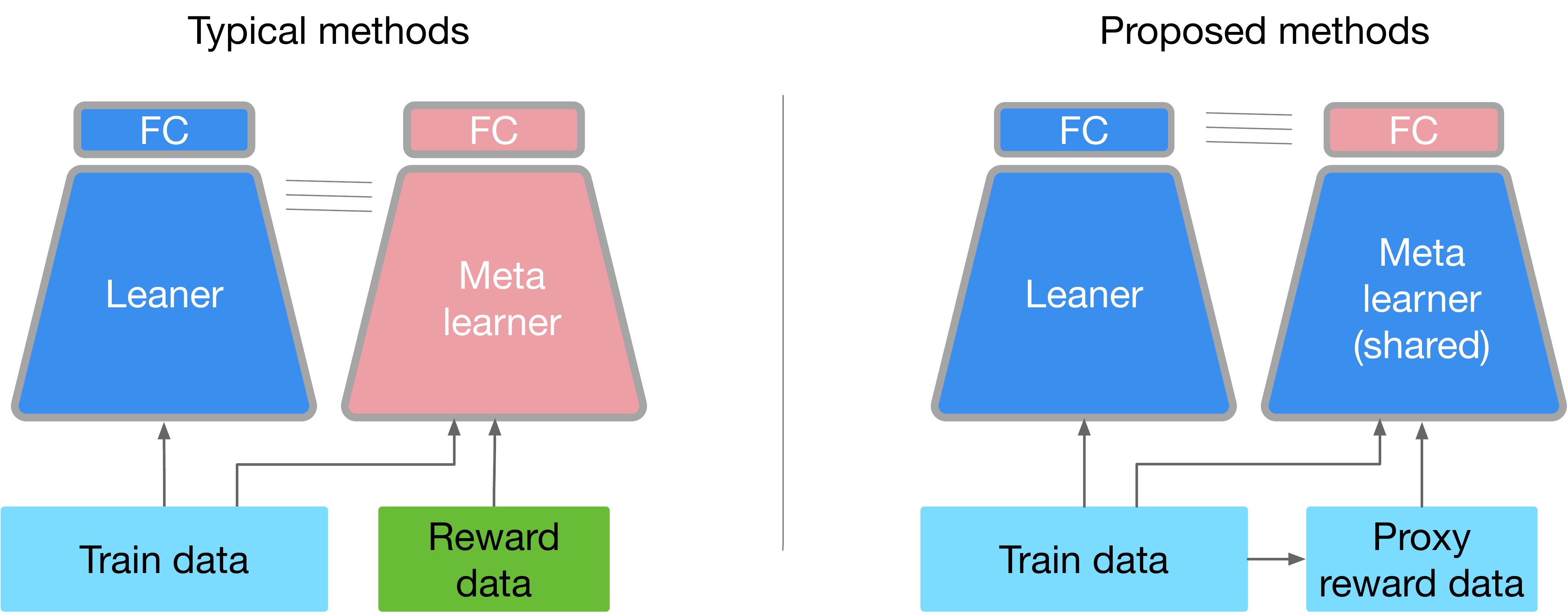}
\caption{Overview of the proposed method compared with typical method for meta re-weighting. Our method is fast and does not need reward data.}
     \label{fig:outline}
     \vspace{-.2cm}
\end{figure}

In this paper, we present a new fast sample re-weighting (FSR) method to overcome the two aforementioned problems (Figure \ref{fig:outline}): a) removing reward data dependence and b) improving training efficiency. To this end we make the following contributions:

\begin{itemize}[itemsep=0.1pt]
    \item We leverage a dictionary (essentially an extra buffer) to monitor the training history reflected by the model updates during meta optimization periodically, and propose a valuation function to discover meaningful samples from training data as the proxy of reward data. 
    The unbiased dictionary keeps being updated and provides reward signals to optimize sample weights. 

    \item Motivated by an investigation we conducted into the mechanism of sample weight meta-objective, instead of maintaining model states for both model and sample weight updates separately, we propose to enable feature sharing for saving the computation cost used for maintaining respective states.
    The proposed method demonstrates significant improvement in training efficiency, which is a desirable feature for large-scale DNN training compared to previous learning-based sample weighting methods.
    
    \item Because our proposed method does not rely on additional reward data, we can directly apply it to tackle common data biases, including noisy labels and long-tailed recognition, as well as more challenging complex of these two types of label corruption.
    Since fast re-weighting ability of FSR is orthogonal to domain-specific techniques, we also propose a momentum re-labeling technique with MixUp regularization to enhance the performance of FSR in noise robust training.  
    Extensive experiments demonstrate our competitive performance compared to previous methods. 
    
\end{itemize}

\section{Related Work}
Training samples are unequally important. 
Weighting training samples is a traditional and effective strategy to improve model performance. Traditional ML methods, such as importance sampling \cite{kahn1953methods}, AdaBoost \cite{freund1997decision}, and self-paced learning \cite{kumar2010self}, explore the usage of important samples for better model training. These weighting strategies have been extended to deep learning \cite{jiang2015self,katharopoulos2018not,liu2015classification,wang2017robust,byrd2019effect}. 
More specialized sample weighting approaches have also been developed in order to weight training samples targeting different goals, either to enhance high value data points, or to reduce data biases (e.g., label noises) \cite{dong2017class,khan2017cost}. 
Hard sample mining \cite{shrivastava2016training} seeks for hard objects to bootstrap model training. 
Focal Loss \cite{lin2017focal} is widely used to allow DNNs to focus on hard samples.
\cite{cui2019class} proposes a class balanced (CB) loss for long-tailed recognition. \cite{saxena2019data} utilizes the softmax temperature to learn instance and class weights.
Moreover, learning based re-weighting methods becomes popular for label noise-robust training. 
For example, \cite{jiang2017mentornet} proposes to train a sequence model to predict good sample weighting. 
\cite{ren2018learning,shu2019meta,hu2019learning} propose meta learning based re-weighting. 
Besides learning sample weights alone, followup methods have explored learning additional data related coefficients, such as labels \cite{zhang2020distilling,pham2020meta,li2019learning} and data augmentation policies \cite{hu2019learning}.

Most of the popular learning based weighting methods originated from meta learning (or learning-to-learn) \cite{hospedales2020meta}, e.g. MAML \cite{finn2017model}. 
A common requirement is the access to a reward dataset that needs to be unbiased, so a meta-objective can be defined to optimize the sample weights. 
This reward dataset is mandatory for conventional meta learning used for few shot learning which requires models to quickly adapt to new tasks. 
However, in the task of learning sample weights, where reward data is drawn from the same distribution of training data, we question and investigate the necessity of that.
Moreover, this type of methods consumes extensive compute and memory. 
Recently, the meta learning community has explored how to accelerate training of MAML \cite{raghu2019rapid,nichol2018first,zintgraf2018caml}. 
However, to our best knowledge, no previous work has studied how to accelerate the training process for learning sample weights.

\section{Background: Learning Sample Weights}
In this section, we briefly review the background of learning based sample re-weighting methods \cite{ren2018learning,shu2019meta,hu2019learning,dehghani2017learning}.

We define a dataset $D = \{(x^D_i, y^D_i), 1 \leq i \leq N \}$ with totally $N$ samples and label $y$ contains a certain degree of biases. Assuming we have another unbiased reward dataset $R = \{(x^R_i, y^R_i), 1 \leq i \leq M \}$ with totally $M$ samples (where $M \ll N$).
The objective of training DNN parameters $\Theta$ can be formulated as a weighted cross-entropy softmax loss,
\vspace{-.2cm}
\begin{equation}
\Theta^{\star}(\mathbf{\omega}) = \arg \min_{\Theta} \sum_{i=1}^N \mathbf{\omega}_i L\big(y^D_i, f(x^D_i; \Theta)\big),
\label{eq:weightobj}
\vspace{-.2cm}
\end{equation}
where $\mathbf{\omega}_i$ is the sample loss weight for $x^D_i$.
$f(\cdot; \Theta)$ is the targeting DNN that outputs class logits and $L(\cdot, \cdot)$ is the standard softmax cross-entropy loss for each training data pair $(x, y)$. In regular supervised training, $\omega$ is equally distributed to samples in each mini-batch. 

Here $\mathbf{\omega}$ is treated as learnable parameters for optimization. To this end, the meta learning is formulated to learn the optimal $\mathbf{\omega}$ for each training data in $D$, such that the trained model with new sample weights can perform best on the reward data in $R$, measured as the reward signal by a cross-entropy loss
\begin{equation}
\begin{split}
\mathbf{\omega}^{\star} = \arg \min_{\mathbf{\omega}, \mathbf{\omega}\ge0} \frac{1}{M} \sum_{i}^{M} L\big(y^R_i, f(x^R_i; \Theta^{\star}(\mathbf{\omega}))\big). \label{eq:weight}
\end{split}
\end{equation}
The problem can be solved by enumerating real value sample weights in a brute-force fashion and training the model until converge at each weight combination. However, the computation requirement is infinite. 
Currently methods that try to address this problem borrow the concept of meta learning to perform a single step model gradient update to online estimating $\Theta^{\star}$: $\Theta_{t+1}(\mathbf{\omega}) = \Theta_{t} - \eta \nabla_{\Theta} \big(\sum_{i} \mathbf{\omega}_i L(y_i, f(x_i; \Theta_t))\big)$, where $\eta$ is a scalar step size.
This enables differentiability of sample weight variables. Therefore, at each timestamp $t$, we can find the current optimal weight for each sample $\omega_{t,i}^{\star}$ through
\vspace{-.2cm}
\begin{equation}
\omega_{o,i} - \alpha \frac{1}{M} \sum_{i=1}^{M} \left.\frac{\partial }{\partial \omega_{t,i}} L\big(y^R_i, f(x^R_i; \Theta_{t+1}(\mathbf{\omega}))\big)\right|_{\mathbf{\omega}_{t,i}=\mathbf{\omega}_{o}}
\label{eq:weight_diff} 
\end{equation}
where $\alpha$ is the step size and $w_o$ is the initial value. \cite{ren2018learning} resets $w_o = 0$ and calculates a new value every iteration. \cite{shu2019meta} treats it as a trainable variable and updates it using SGD. Lastly, the final weights $\omega^{\star}$ are normalized along mini-batch to satisfy $\sum_i \omega_i = 1$.

\begin{figure}[t]
     \centering
     \includegraphics[width=0.999\linewidth]{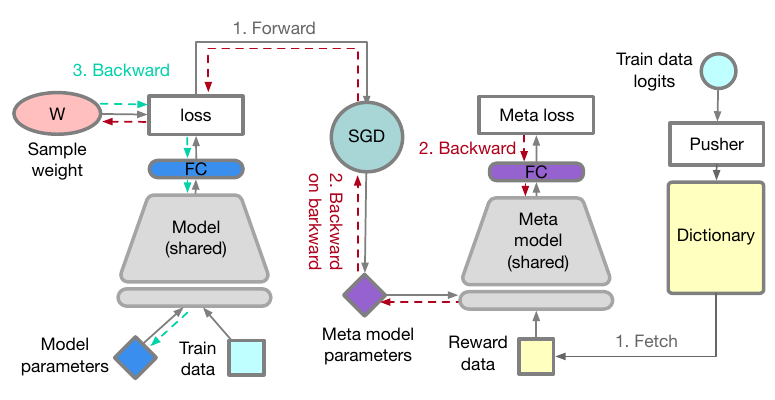}
\caption{Illustration of the three major steps of the method. A dictionary is dynamically updated to maintain the proxy of reward data to enable meta optimization (gradient-by-gradient) of sample weight. Meta model parameters only include partial model parameters (e.g. FC in this example), while the rest are shared. }
     \label{fig:overview}
\end{figure}

\noindent
\textbf{Training Complexity.} 
$\Theta(\mathbf{\omega})$ is a function of $\omega$, the gradient computation of $\mathbf{\omega}$ in Equation \eqref{eq:weight_diff} requires second-order back-propagation (also called gradient-by-gradient in literature). The whole training step requires 1) a forward pass on the training data and 2) reward data once each, 3) a step of gradient descent, 4) second-order back-propagation once, and 5) final model update with SGD. Therefore, there needs $3\times$ training time than regular supervised training as analyzed by \cite{ren2018learning}. 
At the same time, in modern deep learning libraries, the auto-differentiation mechanism will need to hold intermediate representations for second-order back-propagation, which takes higher GPU memory. The experimental section conducts detailed analysis.

\begin{algorithm}[tb]
\small
   \caption{\small FSR training step. The dictionary is updated every epoch. $\Theta^{\text{meta}}$ is pre-defined based on DNN architectures. }
   \label{alg:step}
\begin{algorithmic}
   \STATE {\bfseries Input:} Training data D, model parameter $\Theta$, batch size b, dictionary $R$, reward batch size q, warm-up epoch $E$.
    \STATE Initialize $R_0 = \text{BalancedSampleBatch}(D, |R|)$.
    \FOR{$t=1$ {\bfseries to} $T - 1$}
       \STATE $\{X^D, Y^D\} \leftarrow \text{SampleBatch}(D, b)$.
       
       \IF{$\text{Epoch}(t) \ge E$}
        \STATE $\Theta^{\text{meta}}_{t} \leftarrow \text{Synchronize with } \Theta_{t}$.
        \STATE Initialize $\omega_o \leftarrow \frac{1}{b}$.
        \STATE $\Theta^{\text{meta}}_{t+1} = \Theta^{\text{meta}}_{t} - \alpha \nabla_{\Theta^{\text{meta}}} \sum_{i}^{|X^D|} \omega_i L\big(y_i, f(x_i; \Theta_t)\big)$.
        \STATE $\{X^R, Y^R\} \leftarrow \text{FetchBalancedBatch}(R_t, q)$.
\STATE $\omega^{\star} \leftarrow$ Update sample weights (Equation \ref{eq:weight}).
        \STATE $\omega^{\star} \leftarrow$ Normalize($\omega^{\star}$).
       \ELSE
        \STATE $\omega^{\star} \leftarrow \omega_o$
       \ENDIF
       \STATE $\Theta_{t+1} \leftarrow \text{Update model} \text{ given } \omega^{\star}$ (Equation \ref{eq:weightobj}).
       \STATE $\mathcal{P}(x, \Theta_{t+1}) \leftarrow \text{Compute scores for } x \in X^D$ (Equation \ref{eq:mega-margin}).
       \STATE $\mathcal{\hat{P}}_t(x) \leftarrow \lambda \mathcal{\hat{P}}_t(x) + (1 - \lambda) \mathcal{P}(x, \Theta_{t+1}), \, x \in X^D$.
       \IF{$\text{EpochEnd}(t)$}
        \STATE $R_{t+1} \leftarrow \text{Update dictionary using } \mathcal{\hat{P}}_t(D)$ (Equation \ref{eq:pusher}).  
        \ELSE
          \STATE $R_{t+1} \leftarrow R_{t}$
       \ENDIF
  \ENDFOR
\end{algorithmic}
\end{algorithm}

\section{Method}
This section introduces the proposed method. Algorithm~\ref{alg:step} presents the complete training details of FSR.

\subsection{Learning from Past as Dictionary Fetching}
\label{sec:past}
Instead of preparing an extra reward set, we propose to use a dictionary to store valuable training samples that can be used as a proxy of unbiased reward data, where data biases are controllable since labels and model predictions in $R$ are known. 
The dictionary is dynamically updated to improve its quality.
To this end, we maximize a defined a pusher function $\mathcal{P}$ such that the selected dictionary is
\begin{equation}
R^{\star} = \arg \max_{R} \sum_{x \in R,R \subset D} \mathcal{P}(x, \Theta),
\label{eq:pusher}
\end{equation}
where $R$ has a fixed buffer size as an hyper-parameter. See experiments for more discussions about the memorization assumption behind the definition.

Choosing an effective $\mathcal{P}$ is important for the quality of reward data. This problem connects to an active research field on data valuation \cite{ghorbani2019data}. 
Different from these methods, the valuation calculation here needs to be efficient in order to execute with model updating at every step. We propose the following \emph{meta-margin} definition
\begin{equation}
    \label{eq:mega-margin}
    \mathcal{P}(x, \Theta_t) = L\big(f(x, \Theta_t), y\big) - L\big(f(x, \Theta_{t}^{\star}), y\big),
\end{equation}
where $\Theta_{t}^{\star}$ represents model state after meta update at timestamp $t$. The proposed meta-margin utilizes the states between the model and the meta model on training data.
Maximizing meta-margin intuitively finds samples whose losses have the largest drop after model gradient descent. It prioritizes training samples which is not well recognized (i.e., $L(f(x, \Theta_t), y)$ is high) but its loss can be well minimized after then. 
In contrast, if a sample has small loss (i.e., $L(f(x, \Theta_t), y)$ is low) already (possibly being memorized), the meta-margin will be small as well. 
More importantly, if the loss cannot be minimized or even increases after update, the pusher function de-prioritizes these samples because they are likely undesirable.

\begin{figure}[t]
     \centering
     \includegraphics[width=.8\linewidth]{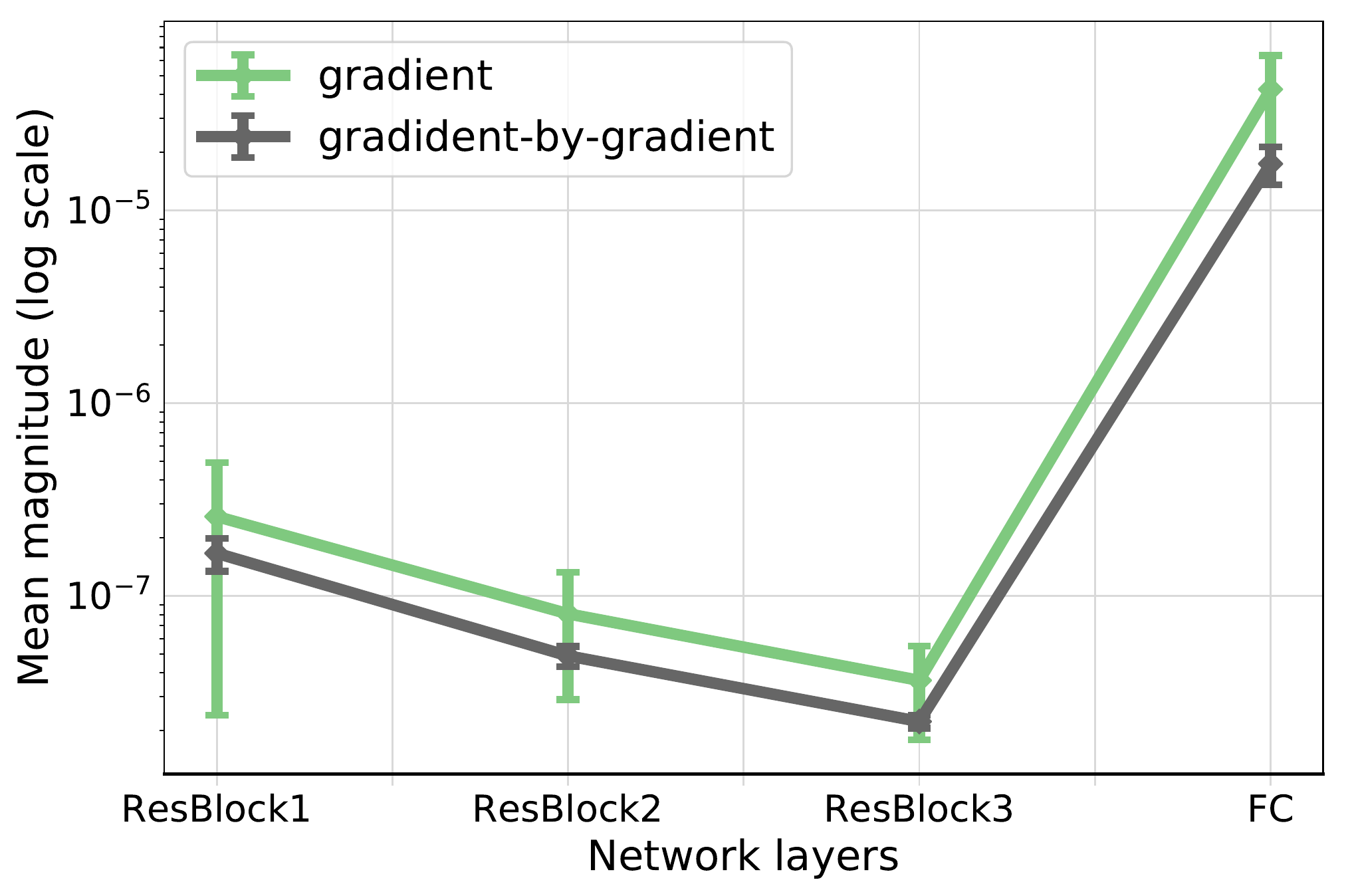}
\caption{Gradient and gradient-by-gradient illustration of Wide-ResNet28-10 residual layer blocks.}
\label{fig:grad_mag}
\end{figure} 

In label noise robust training, why the meta-margin can avoid selecting mislabeled (undesirable) samples? In practice, mislabeled data is usually harder to be minimized than clean data due to the regularization of DNNs that is able to resist label noise. This phenomenon is prominent at early training phases \cite{tanaka2018joint,liu2020early} because models tend to learn simple samples before hard (mislabeled) samples \cite{saxena2019data,bengio2009curriculum}.
It is worth noticing that utilizing this intrinsic regularization to deal with noisy labels is fairly common in previous methods to divide possible clean and noisy samples \cite{pleiss2020identifying,han2018co,liu2020early,ding2018semi}. 
For example, \cite{pleiss2020identifying} proposes to calculate margin between logits of the labeled class and largest logits of other classes to achieve this aim. 
Our proposed meta-margin borrows this high-level inspiration, while at the same time makes use of the meta optimization behaviors to avoid several types of samples that have low value as proxy reward data. 
In experiments, we also explore several existing data valuation candidates as the pusher function to verify the effectiveness of the proposed meta-margin.

\noindent
\textbf{Momentum pusher}. 
The pusher function $\mathcal{P}$ depends on the state of models. To obtain a robust estimation of data points using a series of history of model states, we propose a momentum based pusher function, where $\hat{P}_{t} = \lambda \hat{P}_{t-1} + (1-\lambda) P_{t}$. 
Therefore, we construct $R$ as the reward dataset using the momentum version $\mathcal{P}$ of Equation \eqref{eq:pusher}. 
Lastly, at the start of the training when model is poorly performed, the constructed dictionary could be noisy. Thus we usually setup a couple of epochs for warm-up without applying learned weights (see Algorithm \ref{alg:step} for details).

\subsection{Feature Sharing to Reduce Training Time}
\label{sec:featshare}
Efficiency is an well-known bottleneck for learning based re-weighting, because it requires meta-learning-like nested updates. The experimental section shows quantitive analysis.

We show that improving the training efficiency needs a revisit of model parameter behaviors between nested outer and inner loops.
We speculate the potential possibility of feature sharing between $\Theta_{t+1}$ and $\Theta_{t}$. To be specific, Equation \ref{eq:weight_diff} can be further written as 
\begin{equation}
\small
\label{eq:weight_approx}
\begin{split}
  &  \left.\frac{\partial }{\partial \omega_{t,i}} L^R \right|_{\mathbf{\omega}_{t,i}=\mathbf{\omega}_{o}} \\
  & = \frac{1}{M} \sum_{j=1}^{M} \left.\frac{\partial L_j^R }{\partial \Theta} \right|_{\Theta=\Theta_{t+1}}  \left.\frac{\partial \Theta_{t+1}(\omega)}{\partial \omega_{t,i}} \right|_{\mathbf{\omega}_{t,i}=\mathbf{\omega}_{o}} \\
  & \propto \frac{1}{M} \sum_{j=1}^{M} \sum_{l\in \Theta} \left.\frac{\partial L_j^R }{\partial \Theta_l} \right|_{\Theta_l=\Theta_{t,l}} \left.\frac{\partial L^D_i}{\partial \Theta_l}  \right|_{\Theta=\Theta_{t,l}} \\
  & \propto \frac{1}{M} \sum_{j=1}^{M} \sum_{l\in \Theta^{\text{meta}}} \left.\frac{\partial L_j^R }{\partial \Theta_l} \right|_{\Theta_l=\Theta_{t,l}} \left.\frac{\partial L^D_i}{\partial \Theta_l}  \right|_{\Theta=\Theta_{t,l}}, \\
\end{split}
\end{equation}
where $\Theta_l$ refers to the $l$-th layer of parameters.
The derivation indicates the meta-gradient is the sum of the gradient products of all layers of parameters. 
It motivates us to use partial layers $\Theta^{\text{meta}} \subset \Theta$ to approximate Equation \eqref{eq:weight_approx} (i.e. the last row). To understand the contribution of different layers, Figure \ref{fig:grad_mag} shows the magnitudes of gradient-by-gradient $\nabla_{\Theta_{t+1}} L^R$ and gradient $\nabla_{\Theta_{t}} L^R$ of several layers, suggesting that the changes of sample weights are majorly contributed by the fully-connected (FC) layer. Thus including FC into $\Theta^{\text{meta}}$ could be sufficient to achieve good approximation with the lowest computation cost. 

This approximation is actually equivalent to enabling feature sharing between the model and the meta model, which connects to the recent investigation of rapid learning in MAML \cite{raghu2019rapid}.
In implementation, it is achieved by excluding layers of parameters from bottom up to a certain layer from participating meta optimization (see Figure \ref{fig:overview}). 
For example, if only FC is included, although a reward data mini-batch still needs to forward the meta model, all rest meta gradient-by-gradient computation is reduced to cheap matrix (from FC weights) multiplication.

\subsection{Momentum Re-labeling and Regularization}
\label{sec:relabeling}
Beside re-weighting, we explore more task specific component to improve FSR on noise robust training.
Recent methods favor the design that identifying mislabeled samples and then reusing them as unlabeled data for re-labeling and augmentation \cite{ding2018semi,li2020dividemix}, in light of semi-supervised learning \cite{berthelot2019mixmatch,sohn2020fixmatch,xie2019uda,sohn2020simple,zou2020pseudoseg}. 
FSR intrinsically assigns low weight values to filter mislabeled samples from participating into supervision. 
To enhance FSR for better reuse of those samples, 
we propose a momentum dictionary to maintain long-term prediction estimation of samples and use the estimated predictions as pseudo labels for supervised training. Compared with previous methods with re-labeling, our approach is fairly simpler since it does not need extra copies of extensively augmented data to ensemble pseudo labels or two-stage training to bootstrap mislabeled data \cite{li2017learning,li2020dividemix,ding2018semi}. 

The pseudo label $\hat{y}$ of a sample $x$ at timestamp $t$ is updated in a moving-average manner
\begin{equation}
\hat{y}_t = \beta \hat{y}_{t-1} + (1-\beta) f(x, \Theta_t),
\end{equation} 
where $\beta$ is moving-average decay scalar. 
The batch samples with estimated labels are simply used to construct an extra softmax loss with a multiplier $p$ with the original weighted loss, so the total loss is $\sum_{x_i} \omega_i L(y_i, f(x_i)) + p \cdot L(\hat{y}_i, f(x_i))$.

\noindent
\textbf{Regularization}.
In addition, we apply MixUp \cite{zhang2017mixup} on training data used to compute the weighted loss only (i.e., the first term of the total loss). 
This technique almost becomes a common practice for noise robust methods \cite{song2020learning}. 
In experiments, we find MixUp has strong regularization effects to improve dictionary quality (i.e. label clean ratio) for noise robust training.

\begin{table}[t]
\setlength{\tabcolsep}{5.5pt}
\caption{Test accuracy on CIFAR10 with uniform noise. $^\sharp$ indicates methods requires additional reward data. $\pm$0.0 and `-' mean the corresponding method does not report the value.} \label{tab:cifar10uniform}
\centering
\small 
\begin{tabular}{l|c|ccc}
 \toprule
 \multirow{2}{*}{Method}   & \multicolumn{4}{c}{Noise ratio} \\ \cmidrule{2-5}
                           &    0         & 0.2 & 0.4 & 0.8 \\ \midrule
 GCE      & 93.5 & 89.9$\pm$0.2 & 87.1$\pm$0.2 & 67.9$\pm$0.6 \\
 RoG        &   94.2   & 87.4$\pm$0.0  & 81.8$\pm$0.0 & - \\ 
 MentorNet$^\sharp$    & 96.0 & 92.0$\pm$0.0 & 89.0$\pm$0.0  & 49.0$\pm$0.0 \\
 L2R$^\sharp$       & 96.1 & 90.0$\pm$0.4 & 86.9$\pm$0.2 & 73.0$\pm$0.8 \\ 
 MWN$^\sharp$       &    92.0     &  90.3$\pm$0.6   &  87.5$\pm$0.2  & -   \\  
 CRUST              & 94.4  & 91.1$\pm$0.2 & 89.2$\pm$0.2 & 58.3$\pm$1.8 \\
 ELR                & 94.5 & 92.1$\pm$0.4     &    91.4$\pm$0.2   & 80.7$\pm$0.6     \\ \midrule
 FSR-R32            & 94.4      &    91.8$\pm$0.7       &    90.2$\pm$0.7 & 74.2$\pm$0.9      \\
 FSR                & 96.8      &     \textbf{95.1$\pm$0.1}  & \textbf{93.7$\pm$0.1} & \textbf{82.8$\pm$0.3} \\\bottomrule
\end{tabular}
\vspace{-.3cm}
\end{table}

\begin{table}[t]
\setlength{\tabcolsep}{5.5pt}
\caption{Test accuracy on CIFAR100 with uniform noise. } \label{tab:cifar100uniform}
\centering
\small 
\begin{tabular}{l|c|ccc}
 \toprule
 \multirow{2}{*}{Method}   & \multicolumn{4}{c}{Noise ratio} \\ \cmidrule{2-5}
                           &    0         & 0.2 & 0.4 & 0.8 \\ \midrule
 GCE      & 81.4 & 66.8$\pm$0.4 & 61.8$\pm$0.2 & 47.7$\pm$0.7 \\
 MentorNet$^\sharp$    & 79.0 & 73.0$\pm$0.0 & 68.0$\pm$0.0 & 35.0$\pm$0.0 \\
 L2R$^\sharp$       & 81.2 & 67.1$\pm$0.1 & 61.3+2.0 & 35.1$\pm$1.2 \\ 
 MWN$^\sharp$   &    70.1    &  64.2$\pm$0.3   &  58.6$\pm$0.5  & -   \\
 PENCIL           & 81.4  &  73.9$\pm$0.3 &   69.1$\pm$0.6 & -   \\ 
ELR              & 75.2 & 74.7$\pm$0.3     &    68.4$\pm$0.4   & 30.2$\pm$0.8     \\\midrule
 FSR-R32   &   71.3 &   69.8$\pm$0.2  &  65.9$\pm$0.2 & 41.2$\pm$3.0\\
 FSR   &   81.6      &    \textbf{78.7$\pm$0.2}  & \textbf{74.2$\pm$0.4} & \textbf{46.7$\pm$0.8} \\\bottomrule
\end{tabular}
\vspace{-.5cm}
\end{table}

\begin{table}[t]
\caption{Test accuracy on CIFAR10 with asymmetric noise.} 
\label{tab:cifar10asy}
\centering
\small 
\begin{tabular}{l|cc}
 \toprule
 \multirow{2}{*}{Method}  & \multicolumn{2}{c}{Noise ratio} \\ \cmidrule{2-3}
                          & 0.2 & 0.4 \\ \midrule
 GCE                      & 89.5$\pm$0.3 & 82.3$\pm$0.7 \\
 F-Correction             & 89.1$\pm$0.5 & 83.6$\pm$0.3 \\ 
 PENCIL                   & 92.4$\pm$0.0 & 91.2$\pm$0.0 \\ 
 L2R-R32                  & 89.2 $\pm$0.3  & 84.8$\pm$0.0 \\
 L2R                      & 92.4 $\pm$0.1  & 90.8$\pm$0.3 \\ \midrule
 FSR-R32                  & 91.5$\pm$0.1    & 90.2 $\pm$0.1        \\  
 FSR                      & \textbf{95.0$\pm$0.1}   &  \textbf{93.6$\pm$0.3}       \\  \bottomrule
\end{tabular}
\vspace{-.2cm}
\end{table}

\begin{table}[t!]
    \caption{Top: Hyper-parameters used by different experiments, where most are fixed across experiments. Some hyper-parameters are defined in Algorithm \ref{alg:step}. Bottom: Accuracy with sweeping hyper-parameters on CIFAR100 with $40\%$ uniform noise.} \label{tab:hp}
    \begin{minipage}[b]{0.49\textwidth}
        \setlength{\tabcolsep}{4.0pt}
\small
        \centering
        \begin{tabular}{l|cccccc|cc}
         \toprule
           \multirow{2}{*}{Experiment}  &  \multicolumn{7}{c}{Hyper-parameter} \\
                                        & $|R|$ & $q$ & $b$    & $\lambda$  & $\eta$ & $\alpha$ & $\beta$  & $p$ \\ \midrule
           CIFAR (noise)                & 2k    & 200  & 100   & 0.9        & 0.1 &1 & 0.1 & 2 \\  
           Webvision                    & 5k    & 200  & 16    & 0.9        & 0.1 &1 & 0.1 & 1 \\ \midrule
           CIFAR (Long-Tailed)          & 3k    & 800  & 100   & 0.9        & 0.1 &1 &  - & - \\ 
           iNaturalist2018              & 2k    & 350  & 32    & 0.9        & 0.1 &1 &  - & - \\ \midrule
           CIFAR (LT+noise)             & 3k    & 200  & 128   & 0.9        & 0.1 &1 &  0.1 & 2 \\
           \bottomrule
           
        \end{tabular}
\end{minipage}
\begin{minipage}[b]{0.49\textwidth}
\centering
         \includegraphics[width=.9999\linewidth]{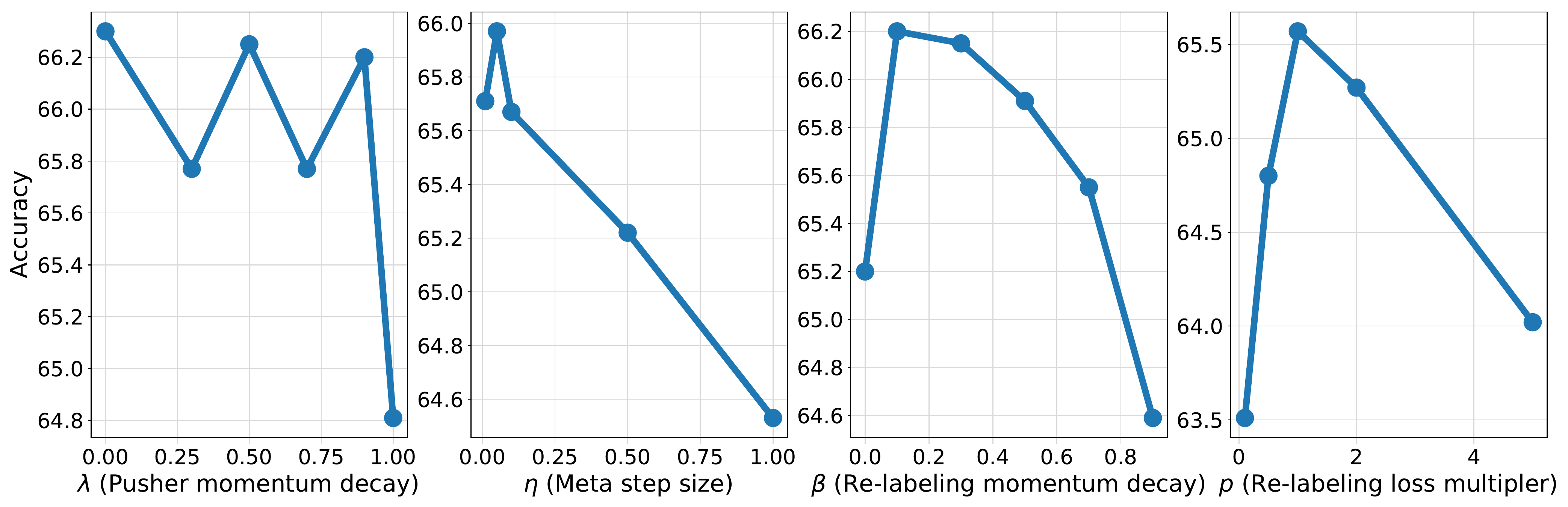}
\label{fig:hyper}
\end{minipage}
    \vspace{-.8cm}
\end{table} 

\begin{table}[t]
 \caption{Test accuracy on WebVision (50 classes). ImageNet validation accuracy on the 50 classes is also reported.} \label{tab:webvision} 
  \centering
  \small 
   \begin{tabular}{l|cc|cc}
    \toprule
     \multirow{2}{*}{Method} & \multicolumn{2}{c}{ImageNet}    & \multicolumn{2}{c}{WebVision} \\ \cmidrule{2-5}
                &   top1        & top5    &   top1        & top5    \\ \midrule  
     F-correction           & 57.4 & 82.4  & 61.1 & 82.7 \\
     D2L & 57.8 & 81.4  &  62.7 & 84.0 \\
     Co-teaching              &  61.5 & 84.7 & 63.6 &  85.2        \\
     Iterative-CV  & 61.6& 85.0 &  65.2 & 85.3 \\
     MentorNet$^\sharp$    & 63.8 & 85.8 & 63.0 & 81.4\\  
     CRUST        &  67.4    &   \textbf{87.8}   &  72.4    &   \textbf{89.6}   \\ \midrule
     FSR          & \textbf{72.3}& 87.2 & \textbf{74.9}  &	88.2  \\\bottomrule
\end{tabular}
\end{table}

\begin{table}[t]
\setlength{\tabcolsep}{4.4pt}
 \caption{Test accuracy on CIFAR long-tailed recognition. FSR results are obtained by averaging 3 runs. DF refers to deferred enabling re-weighting. CB refers to class-balanced loss with different sub-types \cite{cui2019class}. CB-Best adopts the best hyper-parameters for each setup. Note that $\sharp$ indicates methods uses 10 image per class as a reward set, which is not absolute fair comparison.} \label{tab:cifarlongtail}
  \centering
  \small 
   \begin{tabular}{c|ccc|ccc}
    \toprule
     Dataset             & \multicolumn{3}{c}{CIFAR10} & \multicolumn{3}{c}{CIFAR100} \\ \midrule
     Imb. ratio               &  200   & 50  & 10          & 200    & 50     & 10      \\ \midrule  
     SoftMax             & 65.68  & 74.81  & 86.39  & 34.84  & 43.85  & 55.71  \\ 
     CB-Focal            & 65.29  & 76.71  & 86.66    & 32.62  & 44.32  & 55.78  \\
     CB-Best             & 68.89  & 79.27  & 87.49   & 36.23  & 45.32  & 57.99  \\ \midrule  
     L2R$^\sharp$                 & 66.51  & 78.93  & 85.19   & 33.38  & 44.44  & 53.73  \\
     MWN$^\sharp$                 & \textbf{68.91}  & \textbf{80.06}  & 87.84    & \textbf{37.91}  & \textbf{46.74}  & \textbf{58.46} \\\midrule
     FSR-DF              & 66.15  &	79.78  & \textbf{88.15}   &   36.74 & 44.43 & 55.60 \\ 
     FSR                 & 67.76  &	79.17  & 87.40   &	35.44 & 42.57  & 55.45 \\ \bottomrule
\end{tabular}
\end{table}

\section{Experiments}
\subsection{Label Noise Experiments}
We test our method on the CIFAR and WebVision datasets. The compared methods including GCE \cite{zhang2018generalized}, MentorNet \cite{jiang2017mentornet}, RoG \cite{lee2019robust}, L2R  \cite{ren2018learning}, MWN \cite{shu2019meta}, F-correction \cite{patrini2017making}, D2L \cite{ma2018dimensionality}, Co-teaching \cite{han2018co}, Iterative-CV \cite{chen2019understanding}, PENCIL \cite{yi2019probabilistic}, F-Correction \cite{arazo2019unsupervised}, CRUST \cite{mirzasoleiman2020coresets}, and ELR \cite{liu2020early}. 

\textbf{CIFAR.}  We first verify on the common CIFAR10 and CIFAR100 uniform label noises following settings of \cite{ren2018learning,jiang2017mentornet}. We use WRN28-10 as default \cite{zagoruyko2016wide} as it is commonly used in~\cite{ren2018learning,shu2019meta}. We also test using ResNet32 \cite{he2016deep}.
We train the model on a single V100 GPU. We use a cyclical cosine learning rate for $128$ epochs and report the accuracy corresponding to the last iteration. 

Table \ref{tab:cifar10uniform} and Table \ref{tab:cifar100uniform} show the results on the two datasets with different noise ratios, respectively. The proposed FSR outperforms compared methods, and also learning based weight competitors, i.e. L2R, MentorNet, and MWN, which requires additional clean reward data. 
We also verify on asymmetric noise types, which confuses visually similar categories. 
Table \ref{tab:hp} specifies and studies the hyper-parameters used by different experiments. As we can seen, the biggest change by sweeping these hyper-parameters is $\sim2\%$, indicting insensitivity to hyper-parameters. Dictionary size and reward data batch size needs to be changed for different datasets since a class-balanced dictionary size and reward data mini-batch depends to the number of classes.

\textbf{WebVision.} Then, we scale up FSR to large-scale WebVision dataset with 50 classes \cite{jiang2017mentornet}, which contains the top-50 categories of ImageNet. Following the compared methods, ResNet50 is used with random initialization. We use the same training schedule used by CIFAR experiments above. Differently, we train on 8 GPUs for $128$ epochs. 
Table \ref{tab:webvision} compares results to previous methods, demonstrating promising results including the best top-1 accuracy, although the compared CRUST has higher top-5 accuracy.
FSR shows particular strong generalization ability to ImageNet validation accuracy.

\begin{figure*}[t!]
    \begin{minipage}[b]{0.7\textwidth}
         \centering
         \includegraphics[width=.99\linewidth]{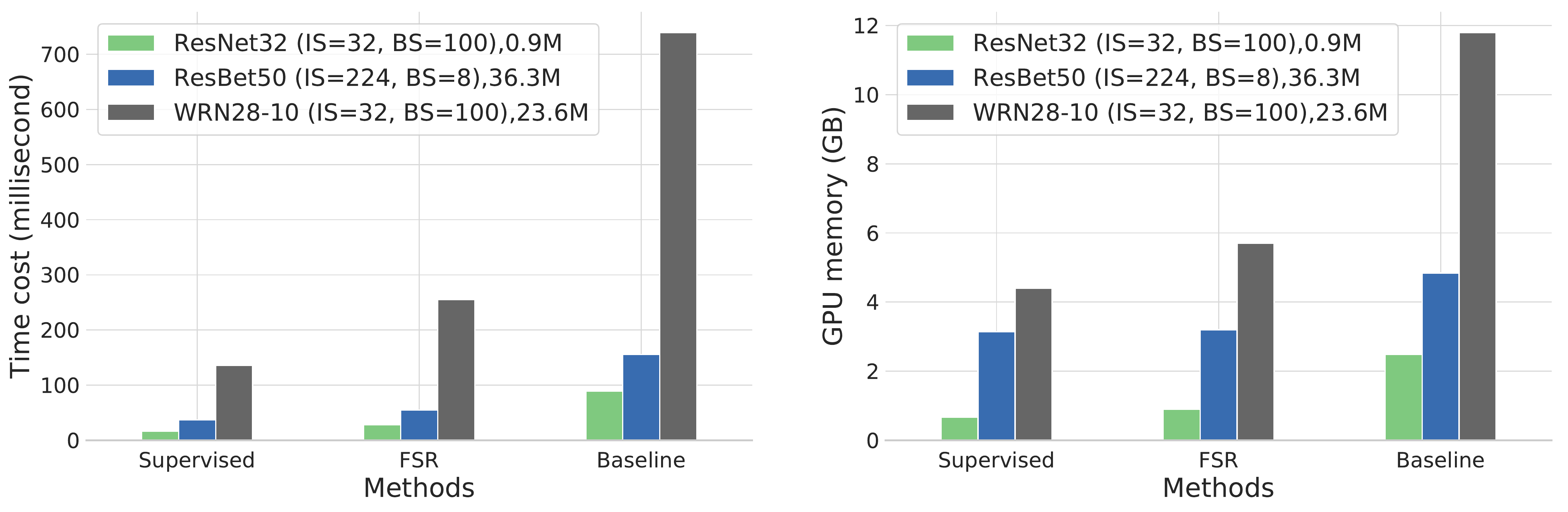}
\end{minipage}
    \begin{minipage}[b]{0.29\textwidth}
         \centering
         \includegraphics[width=.95\linewidth]{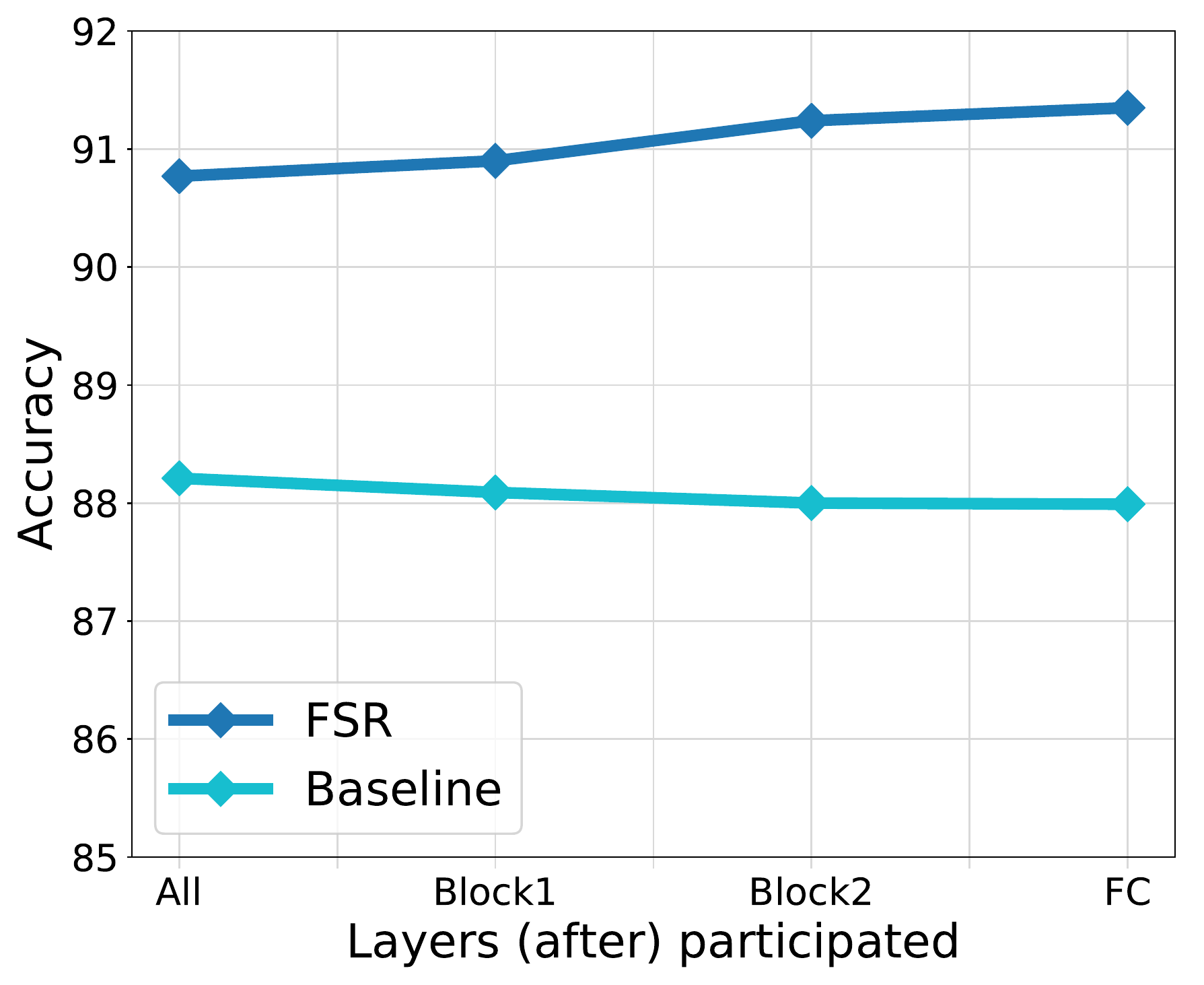}
\end{minipage}
\caption{Left: Training complexity comparison in terms of time per second (left) and GPU memory (right). Three network architectures (with different numbers of parameters) are profiled with certain input image size (IS) and batch size (BS). Right: Accuracy with different numbers of layers included in $\Theta^{\text{meta}}$ for learning sample weights (ResNet32 is used here). $x$-axis indicates the layers after the denoted layer are participated into meta optimization. For instance, FC indicates only the softmax layer is participated (i.e. our default setting)} \label{fig:complexity}
\end{figure*} 
\begin{figure*}[t]
    \small 
    \begin{minipage}[b]{0.33\textwidth}
         \centering
         \includegraphics[width=.99\linewidth]{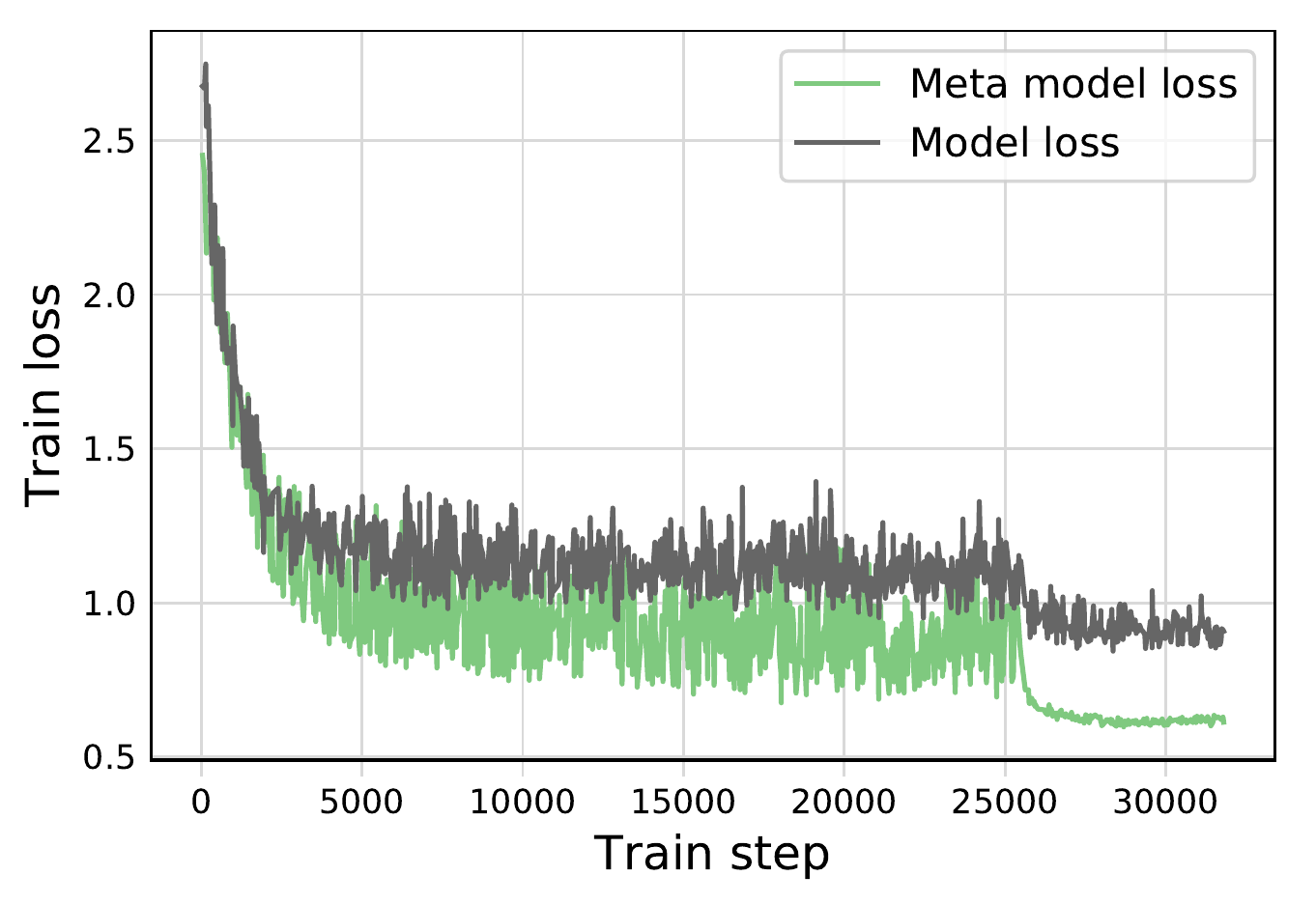}
\end{minipage}
    \hfill
    \begin{minipage}[b]{0.33\textwidth}
        \centering
        \includegraphics[width=.99\linewidth]{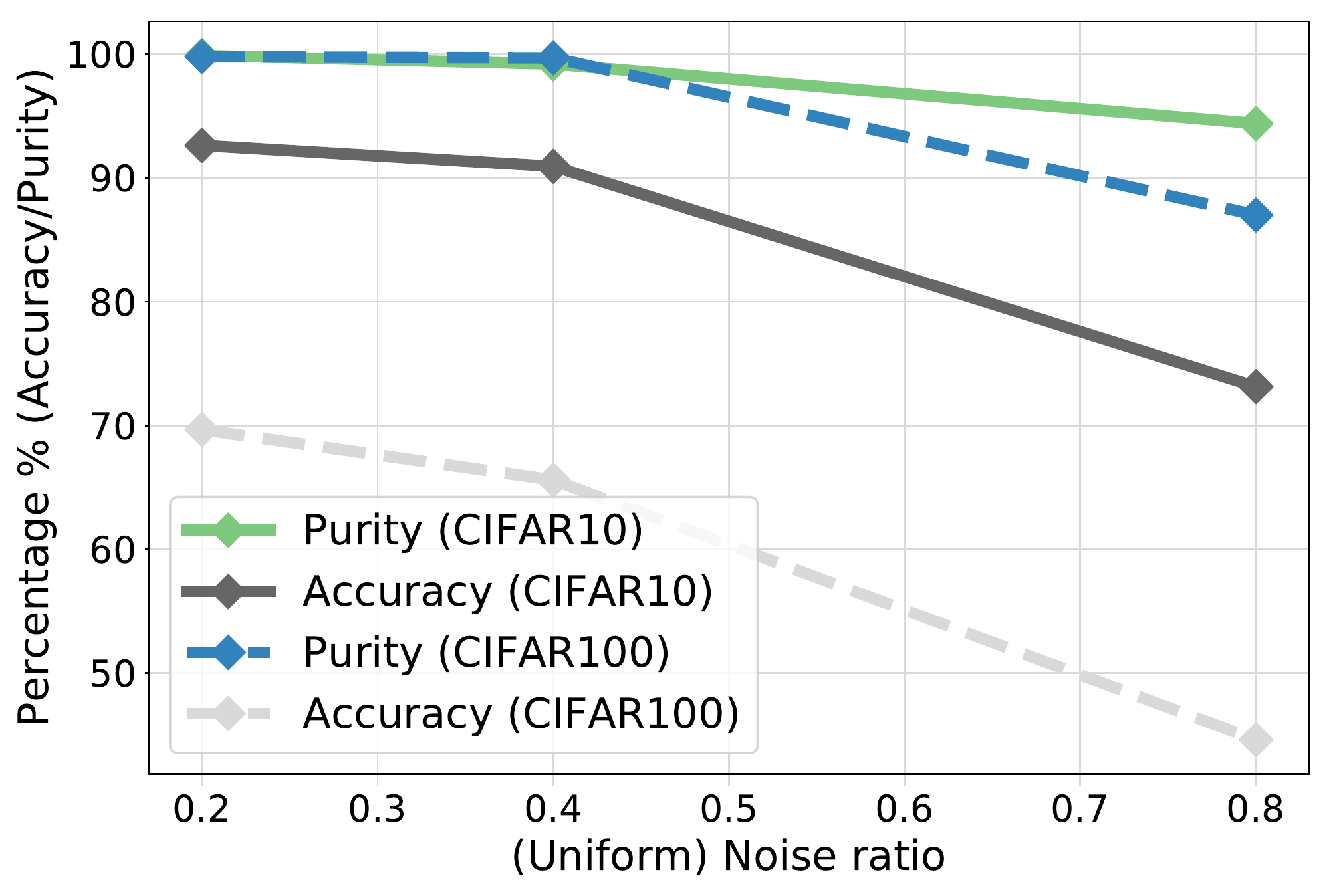}
\end{minipage}
    \hfill
    \begin{minipage}[b]{0.33\textwidth}
        \centering
        \includegraphics[width=.99\linewidth]{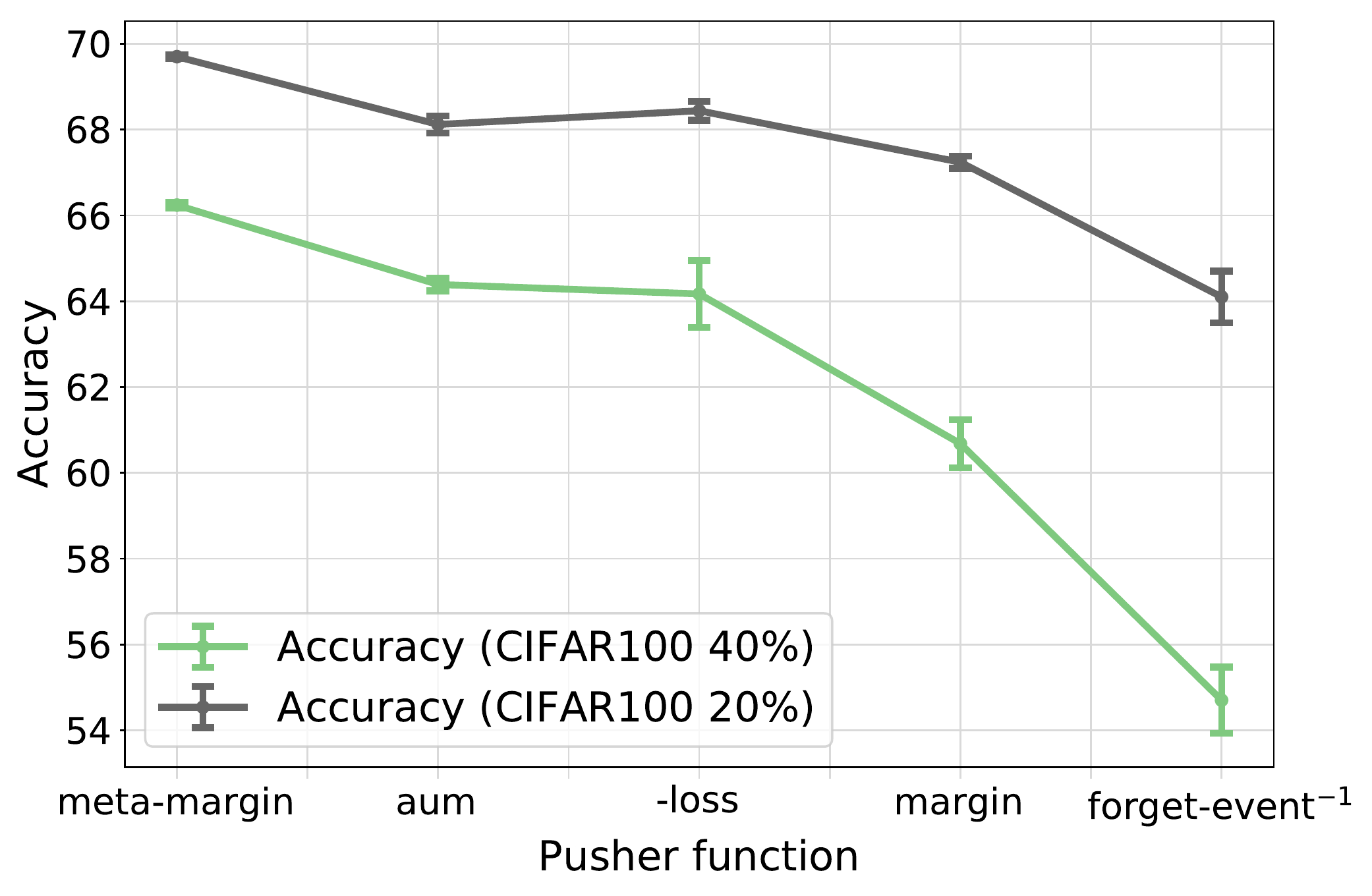}
\end{minipage}
\caption{Ablation study. See explanation in main text. Left: Train loss visualization of model $\Theta_{t}$ and meta model  $\Theta_{t+1}$. Middle: Dictionary purity versus accuracy under different datasets and noises ratios. Right: Comparison of different pusher functions on CIFAR100 with 20\% and 40\% uniform noise.} \label{fig:abla}
\end{figure*}

\subsection{Long-tailed Imbalance Experiments}

\textbf{CIFAR.} We test FSR on long-tailed CIFAR benchmarks, with different imbalance ratios as defined by \cite{cui2019class}.
We modify some training configurations as we use for label noise experiments.
First, we only apply original FSR to test the ability of re-weighting (i.e. without momentum re-labeling and MixUp). 
Second, similar like related methods \cite{cui2019class}, we apply 0.1 softmax label smoothing to compute the loss for re-weighting. Third, Equation \eqref{eq:weight} constraints weight to be non-negative and clips negative values $\text{max}(0, \omega^{\star})$ before normalization. Here we replace this constraint by shifting all weights to be non-zero $(\omega^{\star} - \min(\omega^{\star}) + \frac{1}{b})$ before normalization. 
Fourth, recent methods \cite{cao2019learning,kang2019decoupling} commonly apply deferred balancing, which trains model in the supervised manner to learn representations and then apply re-balancing methods to fine-tune the model. 
We reuse the 200-epoch learning schedule from \cite{cui2019class} (see Table \ref{fig:hyper} for extra hyper-parameter details). 
We optionally apply this deferred schedule by enabling FSR after the first learning rate decay at 160 epoch (i.e., set the warm-up epoch $E=160$). 
We report results with and without the deferred schedule in Table \ref{tab:cifarlongtail}. Overall, FSR achieves competitive results to other methods, although it marginally underperforms MWN, which requires additional balanced reward data. 

\textbf{iNaturalist.} We also test our method on iNaturalist 2018 \cite{iNaturalist2018}, a large-scale long-tailed recognition benchmark. Following the 90 epoch learning schedule of \cite{cui2019class} with deferred schedule, FSR achieves promising results compared to previous methods: $65.52\%$ top-1 ($85.02\%$ top-5) accuracy, against CB Focal \cite{cui2019class} $61.12\% (81.0\%)$ and Remix \cite{chou2020remix} $61.31\%$ ($82.30\%$). 
Considering FSR here is merely a generic re-weighting approach, we believe it can be incorporated with more sophisticated designs for long-tailed recognition to obtain further improvement. 

\begin{table}[t]
\centering
\caption{Results with mixed label corruption in CIFAR10. Uniform noise ratios are added onto different imbalance ratios. \label{tab:mix}} 
    \small 
    \setlength{\tabcolsep}{3.3pt}
	\begin{tabular}{l|cccc|ccccc}
	\toprule
	Noise ratio        &   \multicolumn{4}{c|}{0.2} &   \multicolumn{4}{c}{0.4} \\ \midrule
	Imbal. ratio    &  0      &   10   & 50  & 200  &  0 & 10 & 50  & 200  \\ \midrule
	CRUST \cite{mirzasoleiman2020coresets}             &  90.2  &    65.7 &  41.5 & 34.3   & 89.2 & 59.5 & 32.4 & 28.8 \\
	FSR (Ours)          &   91.8 &    85.7 &  77.4  &  65.5 & 90.2     & 81.6 & 69.8  & 49.5 \\ \bottomrule
	\end{tabular}
	\vspace{-.2cm}
\end{table}

\subsection{Complex of Label Noise and Imbalance} 
Although label noise and imbalance are usually studied as independent research, in real-world applications, these label corruption actually happen simultaneously. This is a more realistic yet challenging setup. We take an initiative to tackle this problem.
To benchmark, we add uniform noise onto imbalanced CIFAR10. Table \ref{tab:mix} compares the results with the best-performing noise-robust method CRUST \cite{mirzasoleiman2020coresets}. 
It can be seen that FSR achieves much higher improvement margin over compared methods on this dataset than that on the noise-robust datasets, suggesting the robustness of FSR to complex label corruption.

\subsection{Training Complexity}
Recent learning based sample weighting methods have similar theoretical training complexity \cite{ren2018learning,hu2019learning,shu2019meta}. We use L2R for direct comparison\footnote{We re-implement the method using the same programming technique for gradient-by-gradient.}. Figure \ref{fig:complexity}(left) compares the training time and GPU memory cost on three architectures\footnote{TensorFlow profiler toolkit \url{https://www.tensorflow.org/guide/profiler}}. Compared to regular supervised training, FSR increases the training cost by $47\% \sim 87\%$ while the baseline increases $319\% \sim 443\%$. 
For GPU memory usage, FSR increases the memory overhead by $1.9\% \sim 34\%$ while the baseline increases $54\% \sim 270\%$.
The results suggest that FSR significantly improves the training efficiency. 
In addition, FSR requires one-stage training only so it is expected to consume less training cost than popular multi-stage methods, such as our compared Co-teaching \cite{han2018co} and Iterative-CV \cite{chen2019understanding}. 

Furthermore, we study how $\Theta^{\text{meta}}$ impacts the final performance. Does strong feature sharing sacrifice accuracy? We control the number of layers included in $\Theta^{\text{meta}}$. Figure \ref{fig:complexity}(right) reports accuracy on CIFAR10 $40\%$ with uniform noise using ResNet32, suggesting that using partial layers does not impact much on either L2R or FSR. 
It is interesting to note that, for FSR, including fewer layers consistently leads to higher accuracy. 
The study indicates a strong feature sharing is feasible between outer loop and inner loop of a meta optimization step.

\subsection{More Studies and Discussions on FSR}
\noindent
\textbf{Meta model memorization.} Memorization is a factor we need to consider to avoid trivial solution.
If all training data are memorized and generates zero training loss, $\mathcal{P}(x, \Theta)$ in Equation \eqref{eq:pusher} will contain no useful information. However, it is unlikely for a well-regularized DNNs to memorize (over-fit) the training dataset. 
Figure \ref{fig:abla}(left) shows the train loss of model $\Theta_t$ and meta model $\Theta^{\star}_t$ at every timestamp on long-tailed CIFAR (imbalance ratio $=10$). Since meta model is a `locally' optimized model, it leads to lower softmax loss in average, yet no phenomenon of over-fitting. This observation also applies to all other datasets and architectures we experimented. 

\noindent
\textbf{Dictionary purity.} In the label noise task, the purity (the clean ratio) of dictionary $R$ plays a critical role in model performance. If undesirable samples are pushed into the dictionary, accuracy can be affected. 
Figure \ref{fig:abla}(middle) visualizes the accuracy versus dictionary purity under different noise ratios. As can been seen, dictionary purity at $80\%$ noise ratio (CIFAR100 and CIFAR100 both) reduces clearly and thereby causes clear accuracy drop.

We further study a variety of pusher function alternatives against the proposed meta-margin. The simplest \emph{negative-loss} prioritizes well-recognized samples. \emph{max-margin} is a popular method in active learning \cite{balcan2007margin} which we use here to select certain samples. \emph{forgetting-event} \cite{toneva2018empirical} finds easy-to-forget samples as they are bad or hard samples. A high forgetting rate indicates corrupted labels.
\emph{area-under-margin} (AUM) \cite{pleiss2020identifying} produces wider margin on clean samples. 
As can been seen in Figure \ref{fig:abla}(right), the proposed meta-margin performs clearly better than negative loss and AUM, though we observe they can all select clean labels with high rates.
Forget-event$^{-1}$ performs the worst. We realize it is negatively impacted by MixUp. 
Removing MixUp will recover CIFAR100 $20\%$/$40\%$ noise to accuracy $68.1\%$/$60.9\%$.

\begin{figure}[t]
     \centering
     \includegraphics[width=.999\linewidth]{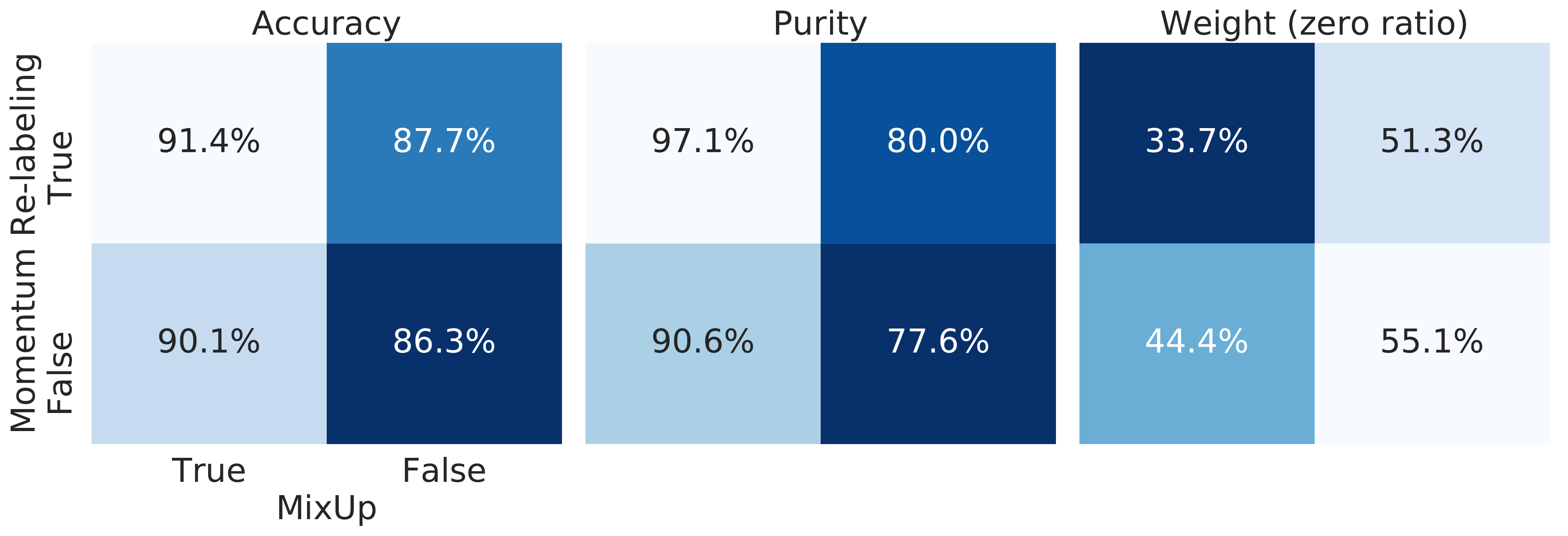}
\caption{Accuracy, dictionary purity, and sample weight zero ratio studies on combinations of momentum re-labeling and MixUp, experimented on CIFAR10 with 20\% uniform noise.}
     \label{fig:purity_vs_mixup}
\end{figure}

\noindent
\textbf{Impact of momentum re-labeling and MixUp.} 
We study how MixUp and the proposed momentum re-labeling  (Section \ref{sec:relabeling}) affect FSR. 
In Figure \ref{fig:purity_vs_mixup}, we show the accuracy, dictionary purity, and also averaged ratio of zero (inactive) weights. 
The ratio of zero weight is expected to be close to the noise ratio of the dataset. Higher ratios of zero weights than actually noise ratios hinder sufficient supervision while lower ratios introduces noisy supervision. 
MixUp has great impact to the dictionary purity, and therefore final accuracy. 
As we can seen in Figure \ref{fig:purity_vs_mixup} (right), the ratio of zero weights is closest to the noise ratio if MixUp is enabled. 
We think MixUp plays a particular regularization role than just augmentation. 
We do not find it is effective for long-tailed recognition. 
Label smoothing \cite{lukasik2020does} is not an effective alternative either. 
We hypothesize it is related to the calibration effects of MixUp \cite{thulasidasan2019mixup} which improves the pusher function and dictionary quality.
Future work is useful for investigation.
Furthermore, momentum re-labeling can further improve all metrics in Figure \ref{fig:purity_vs_mixup}, and pave the last mile to lead accuracy in Table \ref{tab:cifar10uniform}.

\begin{figure}[t]
     \centering
     \includegraphics[width=.8\linewidth]{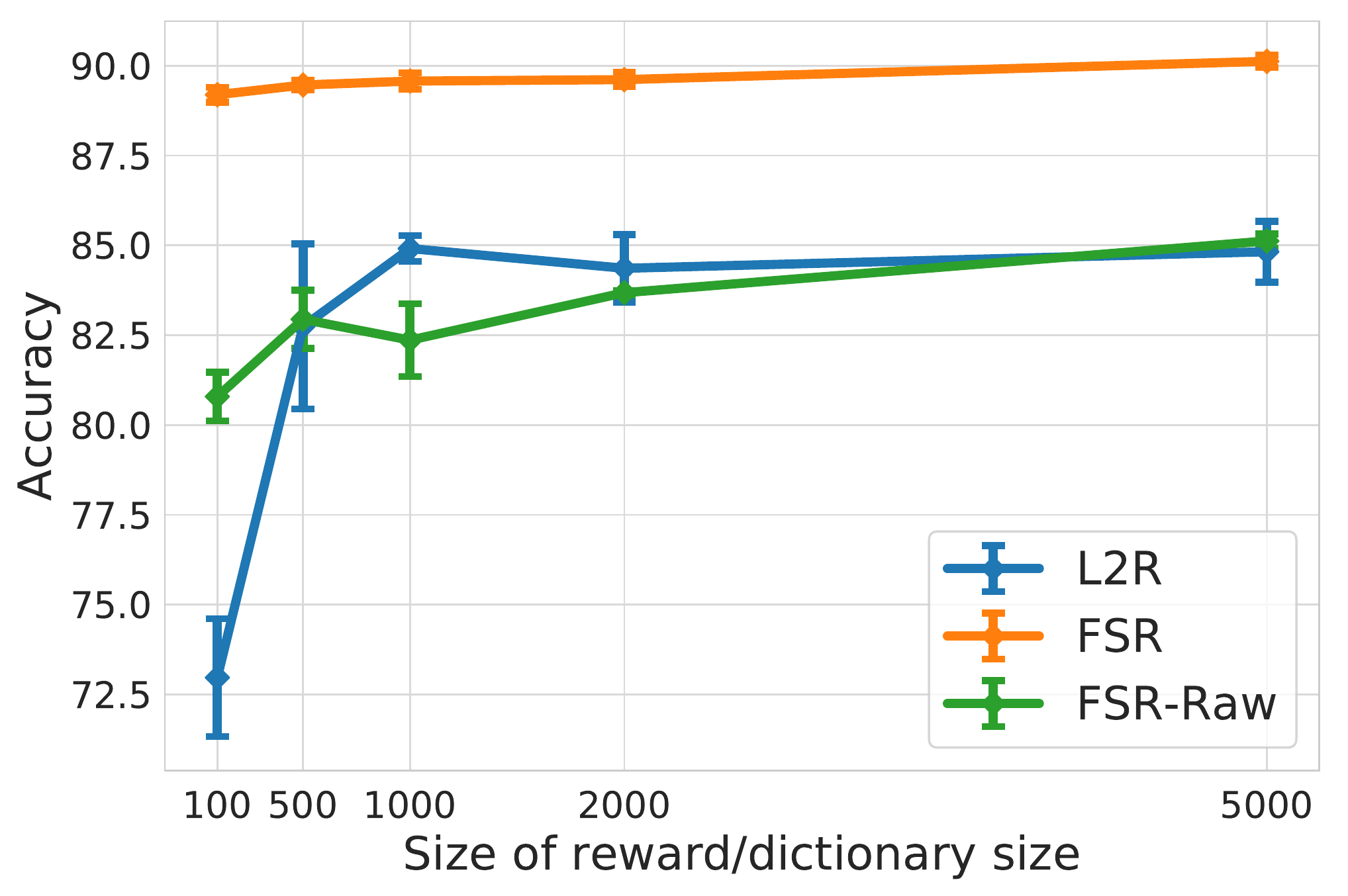}
\caption{Results of the compared L2R, FSR, and FSR-Raw (without momentum re-labeling and MixUp) under varying reward data size (for L2R) and dictionary size (for FSR and FSR-Raw).}
     \label{fig:dicsize}
     \vspace{-.3cm}
\end{figure}

\noindent
\textbf{Reward data versus proxy dictionary.}
\footnote{There are totally 50k training data. In this controllable experiments, a fixed 45k training data is used for all methods. For L2R, the varying reward data is split from the rest 5k data for the different settings. FSR does not use extra data.}
The size of reward data in L2R \cite{ren2018learning} has impacts to the model performance. We train L2R with different reward data size on CIFAR10 with 40\% uniform noise. As shown in Figure \ref{fig:dicsize}, L2R does not benefit much with more reward data and an over-large reward set can even hurt the accuracy. 
The accuracy of L2R also drops largely given very limited reward data. This observation is aligned with the findings by \cite{ren2018learning}.

We then study the impact dictionary size on the accuracy of the proposed FSR.
We conduct the same experiments for FSR and FSR without momentum re-labeling and MixUp (denoted as FSR-Raw). We find FSR is insensitive to the dictionary size and FSR-Raw demonstrates normal sensitivity yet much better than the sensitivity of L2R to real reward data. 
Given sufficient dictionary size, FSR-Raw can perform as good as L2R with arbitrary reward data size. This study further suggests the actual unbiased reward data is dispensable and could be replaced by a well-picked training data subset when with sufficient DNN regularization.

\section{Conclusion}
\vspace{-.1cm}
This paper presents a fast sample re-weighting method, named FSR, that addresses two key bottlenecks for learning based sample weighting methods: high training cost and dependence on additional reward data. The fast re-weighting ability of FSR is orthogonal to domain-specific techniques.
We have shown that by incorporating task specific components (the proposed momentum re-labeling and MixUp), FSR outperforms previous noise robust methods. 
We conduct extensive experiments to demonstrate its effectiveness on noisy labels and long-tailed class recognition benchmarks. 

As future work, we think FSR has the potential to improve other tasks where sample re-weighting matters. 
In addition, we observe from experiments that MixUp has a significant impact on the performance of noisy robust training. Exploring more universal regularization techniques could potentially let FSR generalize better to other domains.

\section*{Acknowledgments}
\vspace{-.1cm}
We would like to thank Chen-Yu Lee, Kihyuk Sohn, and Han Zhang for their valuable discussions.

{\small
\bibliographystyle{ieee_fullname}
\bibliography{egbib}
}

\end{document}